\documentclass[pdflatex,sn-mathphys-num]{sn-jnl}% Math and Physical Sciences Numbered Reference Style
\usepackage{graphicx}%
\usepackage{multirow}%
\usepackage{amsmath,amssymb,amsfonts}%
\usepackage{amsthm}%
\usepackage{mathrsfs}%
\usepackage[title]{appendix}%
\usepackage{xcolor}%
\usepackage{textcomp}%
\usepackage{manyfoot}%
\usepackage{booktabs}%
\usepackage{algorithm}%
\usepackage{algorithmicx}%
\usepackage{algpseudocode}%
\usepackage{listings}%
\theoremstyle{thmstyleone}%
\theoremstyle{thmstyletwo}%

\theoremstyle{thmstylethree}%

\usepackage{makecell}

\newcommand{\ourslong}{Sum-of-Checks}
\newcommand{\OURS}{\ourslong{}}

\newcommand{\datasetendoscapes}{Endoscapes2023}

\begin{document}

% \title[CVSReasonBench: Evaluating End-to-End Surgical Reasoning on Critical View of Safety]{CVSReasonBench: Evaluating End-to-End Surgical Reasoning on Critical View of Safety}

% \title[Sum-of-Cues: Expert-Aligned Visual Reasoning for Surgical Safety with LVLMs]{Sum-of-Cues: Expert-Aligned Visual Reasoning for Surgical Safety with LVLMs}

% \title[Sum-of-Cues: Expert-Aligned Visual Reasoning for Surgical Safety with LVLMs]{Sum-of-Cues: Expert-Aligned Visual Reasoning for Surgical Safety with Large Vision-Language Models}

\title[Sum-of-Checks: Structured Reasoning for Surgical Safety with Large Vision-Language Models]{Sum-of-Checks: Structured Reasoning for Surgical Safety with Large Vision-Language Models}


\author*[1]{\fnm{Weiqiu} \sur{You}}\email{weiqiuy@seas.upenn.edu}

\author[1]{\fnm{Cassandra} \sur{Goldberg}}
\email{cgoldber@seas.upenn.edu}

\author[2,3]{\fnm{Amin} \sur{Madani}}
\email{amin.madani@uhn.ca}

\author[1,4]{\fnm{Daniel A.} \sur{Hashimoto}}
\email{daniel.hashimoto@pennmedicine.upenn.edu}

\author[1]{\fnm{Eric} \sur{Wong}}
\email{exwong@seas.upenn.edu}

\affil[1]{\orgdiv{Department of Computer and Information Science}, \orgname{University of Pennsylvania}, \orgaddress{\city{Philadelphia}, \state{PA}, \country{USA}}}

\affil[2]{\orgdiv{Department of Surgery}, \orgname{University of Toronto}, \orgaddress{\city{Toronto}, \state{ON}, \country{Canada}}}

\affil[3]{\orgdiv{Surgical Artificial Intelligence Research Academy}, \orgname{University Health Network}, \orgaddress{\city{Toronto}, \state{ON}, \country{Canada}}}

\affil[4]{\orgdiv{Department of Surgery, Perelman School of Medicine}, \orgname{University of Pennsylvania}, \orgaddress{\city{Philadelphia}, \state{PA}, \country{USA}}}

% \author*[1,2]{\fnm{First} \sur{Author}}\email{iauthor@gmail.com}



% \affil*[1]{\orgdiv{Department}, \orgname{Organization}, \orgaddress{\street{Street}, \city{City}, \postcode{100190}, \state{State}, \country{Country}}}

% \affil[2]{\orgdiv{Department}, \orgname{Organization}, \orgaddress{\street{Street}, \city{City}, \postcode{10587}, \state{State}, \country{Country}}}

% \affil[3]{\orgdiv{Department}, \orgname{Organization}, \orgaddress{\street{Street}, \city{City}, \postcode{610101}, \state{State}, \country{Country}}}


\abstract{
% \textbf{Purpose:}
% Accurate assessment of the Critical View of Safety (CVS) during laparoscopic cholecystectomy is essential to prevent bile duct injury. While large vision--language models (LVLMs) offer flexible reasoning, their predictions remain difficult to audit and unreliable on safety-critical surgical tasks.
\textbf{Purpose:} Accurate assessment of the Critical View of Safety (CVS) during laparoscopic cholecystectomy is essential to prevent bile duct injury, a complication associated with significant morbidity and mortality. While large vision-language models (LVLMs) offer flexible reasoning, their predictions remain difficult to audit and unreliable on safety-critical surgical tasks.

\noindent\textbf{Methods:}
We introduce Sum-of-Checks, a framework that decomposes each CVS criterion into expert-defined reasoning checks reflecting clinically relevant visual evidence. Given a laparoscopic frame, an LVLM evaluates each check, producing a binary judgment and justification. Criterion-level scores are computed via fixed, weighted aggregation of check outcomes. We evaluate on the Endoscapes2023 benchmark using three frontier LVLMs, comparing against direct prompting, chain-of-thought, and sub-question decomposition, each with and without few-shot examples.

\noindent\textbf{Results:}
Sum-of-Checks improves average frame-level mean average precision by 12--14\% relative to the best baseline across all three models and criteria. Analysis of individual checks reveals that LVLMs are reliable on observational checks (e.g., visibility, tool obstruction) but show substantial variability on decision-critical anatomical evidence.

\noindent\textbf{Conclusion:}
Structuring surgical reasoning into expert-aligned verification checks improves both accuracy and transparency of LVLM-based CVS assessment, demonstrating that explicitly separating evidence elicitation from decision-making is critical for reliable and auditable surgical AI systems.
}




\maketitle

% \linenumbers

\section{Introduction}
\label{sec:intro}

Machine learning models are increasingly deployed in high-stakes medical domains, where incorrect predictions can lead to irreversible harm~\citep{Zarghami2024role,arjmandnia2024value}. However, existing specialized systems and language models typically require substantial task-specific labeled data~\citep{mascagni2025endoscapes,madani2022intraop,sharma2023surgicalactiontripletdetection,seenivasan2023surgicalgpt,wang2025endochat}, limiting their ability to adapt to new clinical questions or changing safety requirements. Large vision-language foundation models (LVLMs) offer a more flexible alternative, leveraging large-scale pretraining to acquire medical knowledge and generalize across tasks~\citep{stueker2025vision,hager2024evaluation}. By jointly reasoning over both visual and textual information, they can support complex clinical prompts beyond single-label prediction.

However, large foundation models still struggle with certain clinically important tasks, particularly those requiring reasoning over diverse and interdependent visual evidence rather than simple recognition of isolated cues~\citep{hager2024evaluation,stueker2025vision}.
One such task is assessing the Critical View of Safety (CVS) during laparoscopic cholecystectomy~\citep{strasberg1995analysis}, which demands jointly evaluating anatomical visibility, spatial configuration, and the explicit exclusion of ambiguity~\citep{strasberg1995analysis,mascagni2022artificial,madani2022intraop,mascagni2025endoscapes,alapatt2025sagescriticalviewsafety}.
% Recent work has fine-tuned LVLMs for surgical understanding and benchmarking~\citep{seenivasan2023surgicalgpt,wang2025endochat}, but these approaches typically require substantial task-specific data. 
Even when augmented with prompting strategies such as chain-of-thought, the resulting reasoning processes are difficult to systematically audit or constrain~\citep{jacovi2020towards,lyu2024towards}, limiting clinicians’ ability to verify whether safety-critical evidence has been explicitly considered.

\begin{figure}[t]
    \centering
    \includegraphics[width=\linewidth]{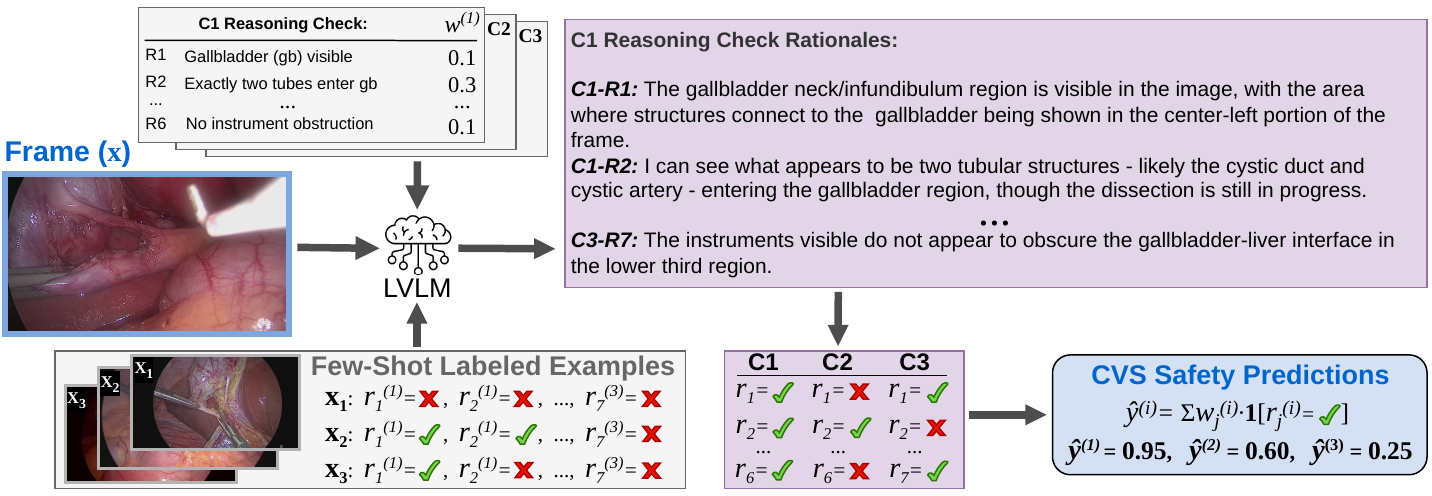}
    \caption{\textbf{Pipeline for structured surgical reasoning with \ourslong{}.} For each CVS criterion, we define a set of expert-aligned reasoning checks, each assigned a fixed weight reflecting its relative importance. Given a laparoscopic frame $x$, we prompt an LVLM to evaluate each check, producing a short justification and a binary judgment. Few-shot examples with labeled reasoning checks are also provided in the prompt context. Criterion-level scores $\hat{y}^{(i)}$ are then computed by aggregating the binarized check outcomes using a weighted sum.}
    \label{fig:main_fig}
\end{figure}

To address this gap, we propose \textbf{\ourslong{}}, a structured framework for surgical safety assessment that decomposes each safety criterion into expert-defined verification checks, shown in Figure~\ref{fig:main_fig}. 
We collaborate with surgeons to define clinically meaningful checks that reflect the evidence they routinely seek (e.g., anatomy visibility, spatial relationships, and observational reliability), and aggregate these check-level judgments into criterion-level scores using a fixed, interpretable weighting scheme. By explicitly separating evidence elicitation from final scoring, \ourslong{} leverages the medical knowledge encoded in LVLMs to predict structured clinical evidence, yielding more accurate safety assessments while enabling fine-grained inspection and targeted intervention when individual checks fail.

Our contributions are threefold:
\textit{(1)} We develop \textbf{\ourslong{}}, a framework for structured surgical safety prediction that operationalizes expert-defined reasoning checks as explicit intermediate decision units.
\textit{(2)} We show that \textbf{\ourslong{}} consistently outperforms free-form reasoning, demonstrating that, even when supplemented by few-shot examples, explicitly structured decision processes are critical for reliable surgical reasoning.
\textit{(3)} We show that while LVLMs reliably predict observational reliability checks (i.e., camera exposure, instrument obstruction), their performance varies substantially on anatomical evidence checks, highlighting a key limitation in current model reasoning for surgical assessment.\footnote{Code is available at \url{https://github.com/BrachioLab/SumOfChecks}.}

\section{\ourslong{}: Structured Surgical Reasoning}
\label{sec:method}

The Critical View of Safety (CVS) is a surgical standard for preventing bile duct injury during laparoscopic cholecystectomy~\citep{strasberg1995analysis}. 
It requires satisfying three criteria:
\textbf{Criterion 1 (C1)}: Two and only two tubular structures are visible entering the gallbladder.
\textbf{Criterion 2 (C2)}: The hepatocystic triangle is cleared of fat and fibrous tissue.
\textbf{Criterion 3 (C3)}: The lower third of the gallbladder is detached from the liver bed~\citep{mascagni2025endoscapes,alapatt2025sagescriticalviewsafety}.
Each criterion is assessed using visual evidence directly observable in individual laparoscopic frames.

\textbf{Problem setup.}
Let $x$ denote a laparoscopic frame, and let $y^{(i)} \in \{0,1\}$ indicate whether criterion $i \in \{1,2,3\}$ is satisfied in that frame.
% Given an LVLM, let $f$ denote the composition of the base LVLM and a fixed post-processing function, yielding predictions $\hat{y}^{(i)} = f(x)$, in which visual evidence extraction and criterion-level reasoning are entangled within a single model invocation.
Our goal is to predict $\hat{y}^{(i)} \in [0,1]$ for each criterion.

\textbf{Expert-defined reasoning checks.}
Rather than predicting each criterion in a single model call, \ourslong{} introduces an explicit intermediate representation that separates evidence elicitation from final decision-making.
For each criterion $i$, we specify a set of expert-defined reasoning checks $\mathcal{R}^{(i)} = \{ R^{(i)}_j \}_{j=1}^{m^{(i)}}$, where $m^{(i)}$ denotes the number of checks associated with criterion $i$, and each $R^{(i)}_j$ is a natural-language verification condition routinely used by surgeons when assessing CVS.
These checks capture key aspects of expert reasoning, including anatomical visibility, spatial configuration, exclusion of ambiguity, and control of occlusions.

Given a frame $x$ and the full set of reasoning checks for criterion $i$, the LVLM produces binary scores
\begin{equation}
r^{(i)}_j = f_j(x, \mathcal{R}^{(i)}) \in \{0,1\}, \quad j = 1, \ldots, m^{(i)},
\end{equation}
where $f_j$ denotes the composition of the LVLM and a fixed post-processing function that maps the model's textual judgment for check $j$ (e.g., \emph{yes}, \emph{no}, or \emph{uncertain}) to a binary value, with affirmative responses assigned $1$ and all other responses assigned $0$, treating ambiguity as insufficient evidence for safety-critical decisions.

% Given a frame $x$ and a reasoning check $R^{(i)}_j$, the LVLM produces a binary score
% \begin{equation}
% r^{(i)}_j = f(x, R^{(i)}_j) \in \{0,1\},
% \end{equation}
% where $f$ denotes the composition of the LVLM and a fixed post-processing function that maps the model's textual judgment (e.g., \emph{yes}, \emph{no}, or \emph{uncertain}) to a binary value, with affirmative responses assigned $1$ and all other responses assigned $0$, treating ambiguity as insufficient evidence for safety-critical decisions.

\textbf{Check aggregation.}
Criterion-level predictions are obtained by aggregating the scored reasoning checks.
For each criterion $i$, we assign non-negative weights $\mathbf{w}^{(i)} = (w^{(i)}_1, \dots, w^{(i)}_{m^{(i)}})$ such that $\sum_{j=1}^{m^{(i)}} w^{(i)}_j = 1$.
The final prediction is computed as a weighted average:
\begin{equation}
\hat{y}^{(i)} = \sum_{j=1}^{m^{(i)}} w^{(i)}_j \, r^{(i)}_j.
\end{equation}
Both the reasoning checks and their weights are defined in collaboration with expert surgeons.
This aggregation scheme enables transparent inspection of how individual checks contribute to the final judgment.

\begin{table}[t]
\caption{\textbf{Frame-level CVS assessment performance (mAP).} \OURS{} outperforms all baseline methods across three models in predicting all three Critical View of Safety (CVS) criteria on Endoscapes. 
Cells report mean average precision (mAP, \%) as mean $\pm$ standard deviation over 3 independent runs with stochastic LLM decoding (temperature $=0.1$). Best results are bolded.}
\label{tab:frame_map_main}
\centering
\setlength{\tabcolsep}{6pt}
\renewcommand{\arraystretch}{1.05}
\scriptsize
\begin{tabular}{l l cccc}
\toprule
\textbf{Model} & \textbf{Method} & \textbf{C1} & \textbf{C2} & \textbf{C3} & \textbf{Avg} \\
\midrule

\multirow{7}{*}{\makecell{\textbf{GPT-4.1}\\\textbf{mini}}}
& Direct 
& 22.9$_{\pm 0.5}$ 
& 23.9$_{\pm 1.4}$ 
& 42.5$_{\pm 1.2}$ 
& 29.7$_{\pm 0.9}$ \\

& Direct+FS 
& 26.9$_{\pm 0.7}$ 
& 26.8$_{\pm 0.4}$ 
& 42.9$_{\pm 0.8}$ 
& 32.2$_{\pm 0.4}$ \\

& CoT 
& 26.2$_{\pm 0.6}$ 
& 27.7$_{\pm 1.7}$ 
& 43.9$_{\pm 0.8}$ 
& 32.6$_{\pm 0.6}$ \\

& CoT+FS 
& 26.9$_{\pm 1.0}$ 
& 28.2$_{\pm 0.4}$ 
& 44.3$_{\pm 0.9}$ 
& 33.1$_{\pm 0.4}$ \\

& SubQ 
& 22.5$_{\pm 1.2}$ 
& 25.9$_{\pm 0.8}$ 
& 43.4$_{\pm 1.2}$ 
& 30.6$_{\pm 0.8}$ \\

& SubQ+FS 
& 22.1$_{\pm 1.2}$ 
& 21.5$_{\pm 0.2}$ 
& 43.8$_{\pm 0.3}$ 
& 29.1$_{\pm 0.5}$ \\

& \textbf{\OURS{}}  (ours)
& \textbf{33.6$_{\pm 0.8}$} 
& \textbf{29.1$_{\pm 1.0}$} 
& \textbf{47.9$_{\pm 1.1}$} 
& \textbf{36.9$_{\pm 0.5}$} \\
\midrule

\multirow{7}{*}{\makecell{\textbf{Claude}\\\textbf{Haiku 4.5}}}
& Direct 
& 20.7$_{\pm 0.3}$ 
& 20.8$_{\pm 0.2}$ 
& 42.2$_{\pm 0.2}$ 
& 27.9$_{\pm 0.1}$ \\

& Direct+FS 
& 21.1$_{\pm 0.2}$ 
& 21.9$_{\pm 0.7}$ 
& 46.5$_{\pm 0.8}$ 
& 29.8$_{\pm 0.4}$ \\

& CoT 
& 19.7$_{\pm 0.7}$ 
& 21.7$_{\pm 0.8}$ 
& 43.8$_{\pm 1.3}$ 
& 28.4$_{\pm 0.4}$ \\

& CoT+FS 
& 22.5$_{\pm 0.5}$ 
& 23.7$_{\pm 0.2}$ 
& 43.1$_{\pm 0.6}$ 
& 29.7$_{\pm 0.4}$ \\

& SubQ 
& 23.4$_{\pm 0.5}$ 
& 22.5$_{\pm 0.7}$ 
& 43.3$_{\pm 0.3}$ 
& 29.8$_{\pm 0.2}$ \\

& SubQ+FS 
& 22.4$_{\pm 0.3}$ 
& 23.8$_{\pm 0.7}$ 
& 42.5$_{\pm 0.3}$ 
& 29.6$_{\pm 0.4}$ \\

& \textbf{\OURS{}}  (ours)
& \textbf{29.2$_{\pm 0.3}$} 
& \textbf{24.2$_{\pm 0.5}$} 
& \textbf{48.5$_{\pm 0.1}$} 
& \textbf{34.0$_{\pm 0.2}$} \\
\midrule

\multirow{7}{*}{\makecell{\textbf{Claude}\\\textbf{Opus 4.5}}}
& Direct 
& 24.3$_{\pm 0.4}$ 
& 25.4$_{\pm 0.1}$ 
& 42.9$_{\pm 0.5}$ 
& 30.9$_{\pm 0.3}$ \\

& Direct+FS 
& 28.0$_{\pm 0.2}$ 
& 26.7$_{\pm 0.2}$ 
& 43.9$_{\pm 0.7}$ 
& 32.9$_{\pm 0.3}$ \\

& CoT 
& 26.4$_{\pm 0.4}$ 
& 26.6$_{\pm 0.8}$ 
& 42.5$_{\pm 1.0}$ 
& 31.9$_{\pm 0.2}$ \\

& CoT+FS 
& 27.6$_{\pm 0.2}$ 
& 26.0$_{\pm 0.4}$ 
& 43.4$_{\pm 0.1}$ 
& 32.3$_{\pm 0.2}$ \\

& SubQ 
& 24.1$_{\pm 1.9}$ 
& 24.5$_{\pm 0.1}$ 
& 43.0$_{\pm 0.6}$ 
& 30.5$_{\pm 0.8}$ \\

& SubQ+FS 
& 19.2$_{\pm 1.1}$ 
& 19.7$_{\pm 1.6}$ 
& 43.9$_{\pm 0.5}$ 
& 27.6$_{\pm 0.9}$ \\

& \textbf{\OURS{}} (ours)
& \textbf{31.1$_{\pm 0.3}$} 
& \textbf{32.9$_{\pm 0.4}$} 
& \textbf{46.9$_{\pm 0.1}$} 
& \textbf{37.0$_{\pm 0.1}$} \\
\bottomrule
\end{tabular}
\end{table}

\begin{table}[t]
\caption{\textbf{\OURS{} Ablation (GPT-4.1-mini).} Few-shot examples and weighted aggregation strategy (over LLM aggregation) are all necessary components for our method's high mAP (\%). \textbf{Bold} is the best result for each column.}
\label{tab:ablation_rubric}
\centering
\setlength{\tabcolsep}{6pt}
\renewcommand{\arraystretch}{1.05}
\scriptsize
\begin{tabular}{l cccc}
\toprule
\textbf{Method} & \textbf{C1} & \textbf{C2} & \textbf{C3} & \textbf{Avg} \\
\midrule
\OURS{} (no FS) & 23.4$_{\pm 0.3}$ & 23.9$_{\pm 1.4}$ & 44.1$_{\pm 0.5}$ & 30.5$_{\pm 0.4}$ \\
\OURS{} (LLM agg.) & 29.3$_{\pm 2.1}$ & 27.2$_{\pm 0.8}$ & 46.5$_{\pm 0.6}$ & 34.3$_{\pm 0.8}$ \\
\textbf{\OURS{} (Weighted)}  (ours) & \textbf{33.6$_{\pm 0.8}$} & \textbf{29.1$_{\pm 1.0}$} & \textbf{47.9$_{\pm 1.1}$} & \textbf{36.9$_{\pm 0.5}$} \\
\bottomrule
\end{tabular}
\end{table}

\section{Experiments}
\label{sec:experiments}

% \noindent\textbf{Experimental setup.}
% We evaluate \ourslong{} on \datasetendoscapes{}, a benchmark dataset for frame-level Critical View of Safety (CVS) assessment in laparoscopic cholecystectomy.
% All experiments use a randomly selected half of the official test split (791 frames) for \datasetendoscapes{}, when using 4 few-shot examples.
% We evaluate our method on three frontier models spanning different model scales and families: GPT-4.1-mini~\citep{openai2025gpt41}, Claude Haiku 4.5~\citep{anthropic2025claudehaiku45}, and Claude Opus 4.5~\citep{anthropic2025claudeopus45}~\footnote{All models are accessed in February 2026.}.
% We compare \ourslong{} against multiple prompting strategies: \emph{Direct}, \emph{Direct+FS}, \emph{CoT}, \emph{CoT+FS}, \emph{SubQ}, \emph{SubQ+FS}.
% Here CoT (Chain-of-Thought) is freeform reasoning before arriving at the final prediction, SubQ first generates question-answering (QA) pairs on its own for each example, and then separately we have the LLM make the prediction from those QA pairs, and we experiment with adding or not adding few-shot examples for each setting.
% Frame-level mean average precision (mAP) is reported in Table~\ref{tab:frame_map_main}.

\noindent\textbf{Experimental setup.}
We evaluate \ourslong{} on \datasetendoscapes{}~\citep{mascagni2025endoscapes}, a benchmark for frame-level CVS assessment in laparoscopic cholecystectomy, using a randomly selected half of the official test split (791 frames) with 4 few-shot examples.
We test on three frontier models: GPT-4.1-mini~\citep{openai2025gpt41}, Claude Haiku 4.5~\citep{anthropic2025claudehaiku45}, and Claude Opus 4.5~\citep{anthropic2025claudeopus45}\footnote{All models are accessed in February 2026.}.
We compare against: \emph{Direct} prompting, \emph{CoT}~\citep{wei2022chain} (free-form chain-of-thought reasoning), and \emph{SubQ} (model-generated QA pairs used as intermediate reasoning before prediction), each with and without few-shot examples (+FS).
Frame-level mean average precision (mAP) is reported in Table~\ref{tab:frame_map_main}.

\paragraph{\ourslong{} improves CVS assessment compared to unstructured prompting.}
As shown in Table~\ref{tab:frame_map_main}, \ourslong{} consistently outperforms all baseline prompting techniques across all three criteria and all LVLMs. Standard CoT and SubQ provide only marginal or inconsistent improvements over Direct prompting, and adding few-shot examples (+FS) yields limited gains for these methods, suggesting that neither free-form reasoning nor exemplar guidance alone reliably improves CVS assessment. In contrast, \ourslong{} effectively leverages both structured checks and few-shot examples to achieve substantial improvements.

\paragraph{Few-shot exemplars and weighted average are both essential for \OURS{}.}
Table~\ref{tab:ablation_rubric} shows that removing few-shot exemplars leads to a substantial performance drop, indicating that structured checks alone are insufficient---mirroring the behavior of SubQ, where the absence of strong exemplar guidance limits reliable prediction. Replacing weighted aggregation with LLM-based aggregation also degrades performance, suggesting that deterministic weighted aggregation provides more stable integration of check-level signals than an additional LLM reasoning step.

% \paragraph{RQ1: Does \ourslong{} improve CVS assessment compared to unstructured prompting?}
% As shown in Table~\ref{tab:frame_map_main}, \ourslong{} consistently outperforms all baseline prompting techniques across all three criteria and all LVLMs.
% In contrast, standard CoT and CoT+FS provide only marginal or inconsistent improvements over Direct prompting, suggesting that free-form reasoning alone does not reliably improve reasoning about the CVS criteria.

% \paragraph{RQ2: How important are each parts of \OURS{}?}
% Table~\ref{tab:ablation_rubric} shows that removing few-shot examples leads to a substantial performance drop, indicating that structured checks alone are insufficient. This mirrors the behavior of SubQ and SubQ+FS, where the absence of strong exemplar guidance limits reliable perception prediction for dynamically generated questions. 

% We further replace weighted aggregation with LLM-based aggregation and observe a consistent degradation in performance. This suggests that, in our setting, deterministic weighted aggregation provides more stable and reliable integration of check-level signals than relying on an additional LLM reasoning step.

\begin{figure}[t]
    \centering
    \includegraphics[width=\linewidth]{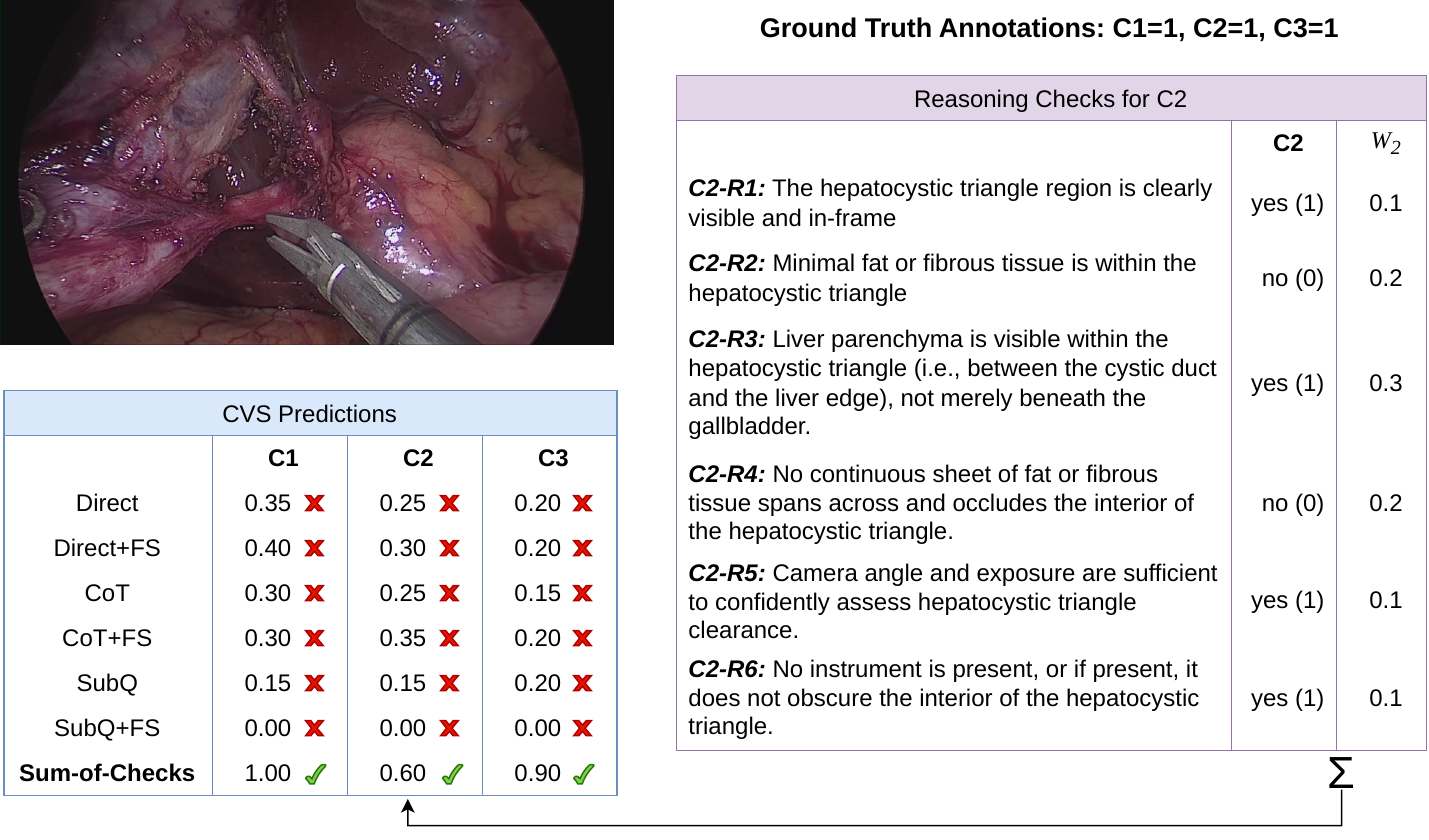}
    \caption{\textbf{Example output of \ourslong{} (Claude Opus 4.5).} 
All three CVS criteria are satisfied (ground truth = (1,1,1)). Standard prompting methods fail on all criteria, whereas \OURS{} correctly predicts all three. For C2 (hepatocystic triangle clearance), the model identifies satisfied sub-checks (e.g., visibility of liver parenchyma, sufficient exposure), yielding a weighted score of 0.6 $>$ 0.5 and a correct prediction.}
    \label{fig:example_output_fig}
\end{figure}

\section{Discussion and Conclusion}
\label{sec:discussion}
\OURS{} substantially improves frame-level CVS assessment by decomposing safety criteria into expert-defined verification checks that explicitly separate evidence elicitation from decision-making, improving both accuracy and interpretability. Figure~\ref{fig:example_output_fig} illustrates this on a representative frame where all baselines fail but \OURS{} correctly predicts all three criteria.

\paragraph{Reasoning check prediction reliabilities.}
LVLMs' performance in completing reasoning checks vary substantially across check types, even within the same criterion. Observational checks---such as camera exposure, field of view, and instrument obstruction---are predicted consistently, while checks requiring anatomical evidence and spatial reasoning exhibit much greater variability. This indicates that current LVLMs remain less reliable on the decision-critical evidence required for complex CVS assessments.

\paragraph{Few-shot exemplar selection.}
Randomly selected exemplars are insufficient for reliable predictions. We select 4 diverse examples with label combinations (000, 111, 110, 001) and find two properties of effective exemplars: (1)~\textit{typicality}---clear, canonical instances of each condition; and (2)~\textit{minimal variation}---paired exemplars differing in only one criterion should share the same visual context. Our results depend on in-distribution exemplars; we leave out-of-distribution cases to future work.

\section*{Limitations}
We evaluate on Endoscapes2023~\citep{mascagni2025endoscapes}, which provides expert-annotated key frames aligned with our frame-level, check-based evaluation protocol. Other datasets such as Cholec80-CVS~\citep{rios2023cholec80} and the SAGES CVS Challenge~\citep{alapatt2025sagescriticalviewsafety} differ in annotation granularity or distribution; we leave these extensions, along with application to other safety-critical reasoning tasks, to future work.

\section*{Declarations}

% Competing Interests: Authors are required to disclose financial or non-financial interests that are directly or indirectly related to the work submitted for publication. Please refer to “Competing Interests and Funding” below for more information on how to complete this section.

\subsection*{Funding}
This research was partially supported by a gift from AWS AI to Penn Engineering's ASSET Center for Trustworthy AI, by ASSET Center Seed Grant, ARPA-H program on Safe and Explainable AI under the award D24AC00253-00, by NSF award CCF 2442421, by the AI2050 program at Schmidt Sciences (Grant G-25-67983), and by funding from the Defense Advanced Research Projects Agency's (DARPA) SciFy program (Agreement No. HR00112520300). The views expressed are those of the author and do not reflect the official policy or position of the Department of Defense or the U.S. Government.
Daniel Hashimoto was partially supported by the American Surgical Association Foundation for this work. 

\subsection*{Conflicts of interest}
Daniel Hashimoto is a consultant for Medtronic.

\subsection*{Ethical approval}
This study used publicly available, fully de-identified surgical video data with no new human data collection or patient interaction; ethical approval was not required.

\subsection*{Informed consent}
Not applicable. The study used publicly available, de-identified data.

\begin{appendices}

\end{appendices}


\bibliography{sn-bibliography}% common bib file

\end{document}